\documentclass[conference]{IEEEtran}

\IEEEoverridecommandlockouts                              

\usepackage{cite}
\usepackage{amsmath,amssymb,amsfonts}

\usepackage{graphicx}
\usepackage[small]{caption} 
\usepackage[caption=false]{subfig}
\usepackage{textcomp}
\usepackage{xcolor}
\usepackage{times}
\usepackage{soul}
\usepackage{url}
\usepackage[hidelinks]{hyperref}
\usepackage[utf8]{inputenc}
\usepackage{algorithm}
\usepackage{algorithmic}
\usepackage{multicol}
\usepackage{multirow}
\usepackage{rotating}

\usepackage{xspace}

\begin{document}

\title{\LARGE \bf
AYDIV: Adaptable Yielding 3D Object Detection
via Integrated Contextual Vision Transformer
}

\author{Tanmoy Dam$^{1}$,  Sanjay Bhargav Dharavath$^{2}$, Sameer Alam$^{1}$, Nimrod Lilith$^{1}$, Supriyo Chakraborty$^{2}$ \\ and Mir Feroskhan$^{1}$ 

\thanks{${1}$ The authors would like to thank the funding support by the Start-Up Grant from the School of Mechanical and Aerospace Engineering at Nanyang Technological University.}

\thanks{${1}$ Tanmoy Dam, Sameer Alam, Nimrod Lilith, and Mir Feroskhan are associated with the Saab-NTU Joint Lab,
        Nanyang Technological University, Singapore
        {\tt\small tanmoydam@yahoo.com, (sameeralam, nimrod.lilith, mir.feroskhan)@ntu.edu.sg}}

\thanks{${2}$ Sanjay Bhargav Dharavath and Supriyo Chakraborty are associated with the Indian Institute of Technology, 
        Kharagpur, India
        {\tt\small (sanjay810, supriyochakraborty)@iitkgp.ac.in}}
}




\maketitle

\begin{abstract}

Combining LiDAR and camera data has shown potential in enhancing short-distance object detection in autonomous driving systems. Yet, the fusion encounters difficulties with extended distance detection due to the contrast between LiDAR's sparse data and the dense resolution of cameras. Besides, discrepancies in the two data representations further complicate fusion methods. We introduce AYDIV, a novel framework integrating a tri-phase alignment process specifically designed to enhance long-distance detection even amidst data discrepancies. AYDIV consists of the Global Contextual Fusion Alignment Transformer (GCFAT), which improves the extraction of camera features and provides a deeper understanding of large-scale patterns; the Sparse Fused Feature Attention (SFFA), which fine-tunes the fusion of LiDAR and camera details; and the Volumetric Grid Attention (VGA) for a comprehensive spatial data fusion. AYDIV's performance on the Waymo Open Dataset (WOD) with an improvement of 1.24\% in mAPH value(L2 difficulty) and the Argoverse2 Dataset with a performance improvement of 7.40\% in AP value demonstrates its efficacy in comparison to other existing fusion-based methods. Our code is publicly available at \textcolor{red}{https://github.com/sanjay-810/AYDIV2} 

\end{abstract}

\begin{IEEEkeywords}
GCFAT, SFFA, VGA, Multi-modal fusion, 3D object detection
\end{IEEEkeywords}


\section{INTRODUCTION}

Automated driving perception systems rely on a range of sensors to continually enhance their performance in critical driving scenarios. Among these sensors, LiDAR and cameras play vital roles in autonomous vehicles, particularly in the domain of 3D object detection (3D OD), which involves localization and classification \cite{arnold2019survey}. LiDAR, despite its importance, provides low-resolution information, numerous methods  \cite{qi2017pointnet, qi2017pointnet++, lang2019pointpillars, wang2021pointaugmenting} have been explored to achieve competitive performance across various benchmark datasets. Nonetheless, owing to the inherent constraints of LiDAR sensors, the point cloud data they generate is typically sparse and lacks the necessary contextual information to effectively differentiate distant areas, ultimately leading to inferior performance \cite{li2022deepfusion}.

Sensor fusion, particularly between LiDAR and camera data, is a complex task due to the challenge of aligning features from both sources. Two main approaches dominate the research: early-stage fusion and mid-level fusion. Early-stage fusion, as exemplified by references like \cite{vora2020pointpainting} and \cite{9578812}, combines the two data sources immediately. On the other hand, mid-level fusion, demonstrated by methods in \cite{huang2020epnet}, \cite{liang2020deep}, and notably by MV3D\cite{chen2017multiview} and AVOD\cite{ku2018joint}, integrates features post extraction. MV3D uses Region of Interest (RoI) fusion for a nuanced combination, while AVOD emphasizes high recall by blending image and Bird's Eye View (BEV) features. The MMF approach \cite{liang2020multitask} integrates 2D detection and depth, enhancing 3D detection accuracy. Achieving accurate data correspondence between sources is pivotal for effective fusion. 

\begin{figure*}
    \centering
    \scalebox{0.28}{
    \includegraphics{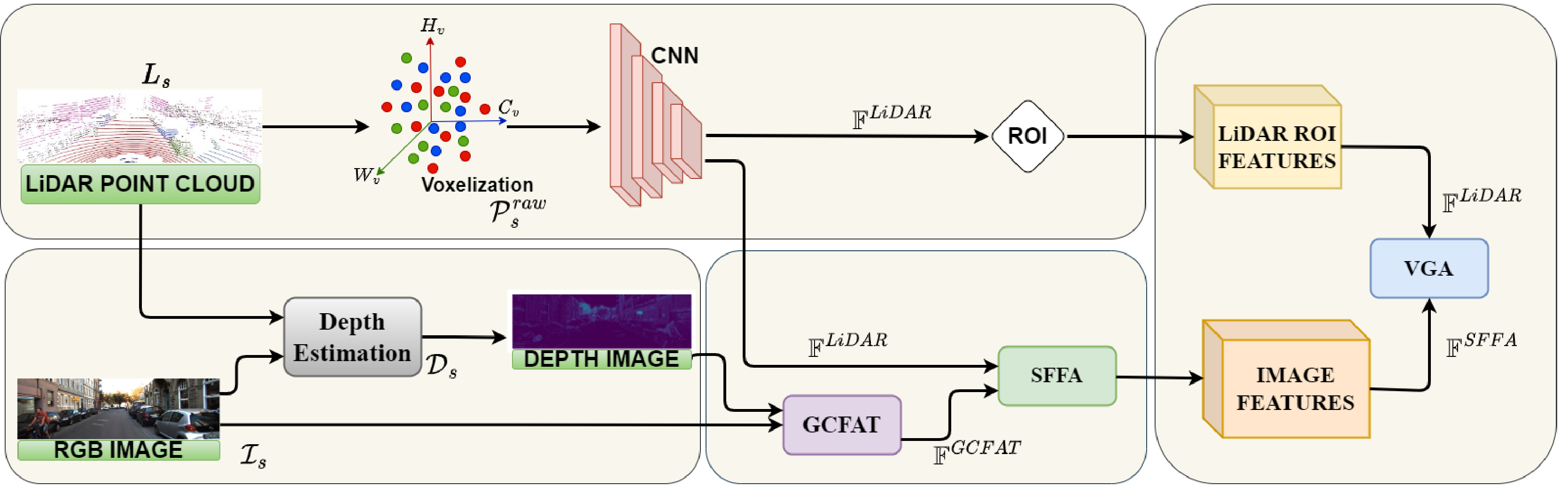}}
    \caption{AYDIV Pipeline: The pipeline integrates multiple modalities through three essential parts: GCFAT, SFFA, and VGA. Each of these parts plays an important role in the complex interaction of local and global contextual attention mechanisms. GCFAT considers the camera as a local feature and the depth information as a global query through its image and LiDAR point clouds. SFFA aligns the image features extracted by GCFAT with the voxelized LiDAR features in a cohesive feature alignment manner. Finally, VGA combines sparse voxelized LiDAR features and SFFA camera features in a grid space.}
    \label{fig:aydiv_main}
\end{figure*}

To address the preceding problems, we propose a novel fusion Network, termed AYDIV, which performs LiDAR-camera fusion at both global and local levels, as shown in Fig \ref{fig:aydiv_main}. Our AYDIV  is
comprised of three novel components: Global Contextual Fusion Alignment Transformer (GCFAT), Sparse Fused Feature Attention (SFFA), and Volumetric Grid Attention (VGA). To offer detailed region-specific data for objects at varying distances while preserving the precise positional information with greater granularity, we propose \textbf{GCFAT} method enhances image feature extraction by merging depth estimation with RGB images, utilizing two attention mechanisms: Local Multi-Scale Attention (LMSA) for small-scale details and Global Diffuse Attention (GDA) for broader patterns, resulting in a comprehensive image understanding. \textbf{SFFA} offers a unique sparse attention mechanism to integrate voxelized LiDAR features data with image features, leveraging the Rectified Linear Unit (ReLU) over the conventional sigmoid function in its attention block, potentially optimizing image recognition. In contrast, \textbf{VGA} focuses on 3D RoI features fusion rather than 2D, offering enriched spatial data with depth details, proving crucial for tasks like 3D object recognition.

Our AYDIV has showcased outstanding performance in 3D object detection, excelling on both the WOD and the AV2. Impressively, AYDIV outperform all current methods that employ both camera and LiDAR for 3D detection on these datasets, achieving an impressive \textit{82.04 mAPH} (L2) detection rate on WOD.

To summarize, our main contributions to this paper are described as follows:
\begin{itemize}
\item We are the first to integrate transformer blocks with the GCFAT structure, enabling the fusion of global depth information with RGB images and thereby enhancing the extraction of depth features from RGB data.
\item Using the SFFA framework, we introduce a method that fuses voxel point cloud and image features through a sparse attention mechanism, optimizing their integration.
\item Our novel RoI feature fusion VGA method refines the fusion process between pseudo point clouds and image features, leading to final integration.
\item Tests on multiclass 3D datasets like Waymo and Argoverse 2 show AYDIV's consistent performance across varied distances, underlining its effectiveness in 3D object detection.
\end{itemize}


\section{RELATED WORK}
\textbf{LiDAR for 3D Object Detection.}
LiDAR point clouds, typically described as disorganized collections of data points, can be broadly categorized into three subgroups: voxel-based, point-based, and point-voxel fusion methods. Voxel-based techniques, as demonstrated in works such as \cite{deng2021voxel,chen2022mppnet,hu2022afdetv2}, group the point cloud data into voxels and subsequently employ deep sparse convolution layers to extract features from these voxels. Point-based approaches, as illustrated in \cite{chen2022mppnet,shi2020point}, involve passing raw point cloud data through stacked Multi-Layer Perceptrons (MLPs) to derive point-level features. Recent studies, as evidenced by \cite{fan2022embracing,vora2020pointpainting}, have embraced a hybrid approach that captures both point and voxel-based representations to obtain a comprehensive feature representation. Alternatively, LiDAR point clouds can also be represented as high-resolution range images, enabling predictions based on depth, as exemplified in \cite{yin2021center,bewley2020range}.



\textbf{LiDAR-camera Integration for 3D Object Detection.}
Combining monocular recognition with LiDAR-generated point clouds enhances 3D object detection \cite{chen2016monocular, tao2023pseudo, ku2019monocular, shi2023multivariate, tao2023weakly}. Monocular systems can predict 3D boxes from 2D images but lack depth information \cite{tao2023pseudo}. To address this, monocular detectors estimate pixel-level depth \cite{tao2023pseudo}. 2D image object recognition serves as a starting point for point cloud data analysis \cite{qi2018frustum, wang2019frustum, liu2023flatformer}, often achieved through a two-step object-centered fusion in previous studies \cite{qi2018frustum, ku2018joint}.

Mid-level Fusion approaches, exemplified by Deep Continuous Fusion \cite{liang2018deep, shi2023multivariate} and others \cite{piergiovanni20214d, zheng2022beyond}, aim to seamlessly integrate 2D and 3D modalities by exchanging information between their backbones. However, the lack of an effective matching mechanism between camera and LiDAR features presents a challenge \cite{li2023logonet}.  Furthermore, aggregating multiple LiDAR points within the same 3D voxel introduces complexities in handling corresponding camera features with varying degrees of importance for 3D detection. In contrast, our fusion matching aggregation network overcomes these challenges by leveraging two distinct modalities, enabling a more robust and accurate representation of the environment and enhancing multi-modal 3D OD performance.  

\section{AYDIV}
The AYDIV pipeline, depicted in Figure \ref{fig:aydiv_main}, revolves around three key components: GCFAT, SFFA, and VGA. For simplicity, we consider the given multi-modal input-output sequences defined by $\{(\mathcal{I}_s,{L}_s), (\mathcal{I}_{(s-1)},{L}_{(s-1)},...\}$, where the $s$-th input sequence comprises two modalities: LiDAR represented as ${L}_s$ and camera image as $x \in \mathcal{I}_s \in \mathbb{R}^{H \times W \times 3}$. The raw LiDAR point cloud of the $s$-th input is represented as $L_{s} \rightarrow \mathcal{P}_{s}^{raw}$ where  
$\mathcal{P}_{s}^{raw} = \{(\mathcal{U}_p, \mathcal{V}_p, \mathcal{W}_p, \mathcal{R}_p)\}_{p=1}^{T}$, where $(\mathcal{U}_p, \mathcal{V}_p, \mathcal{W}_p)$ denotes the position of the LiDAR point, $\mathcal{R}_p$ is the intensity and $T$ is the total number of points.

\textbf{Depth Estimation($\mathcal{D}_s$).} 
Utilizing sparse LiDAR-generated point cloud data($\mathcal{P}_{s}^{raw}$) in conjunction with correlated RGB imagery, denoted as $\mathcal{I}_s$, proves advantageous for extracting globally correlated features within the image feature extraction module. When provided with a set of point clouds represented as $\mathcal{P}_{s}^{raw}$, we have the capability to transform them into a sparse depth map, $\mathcal{D}_s \in \mathbb{R}^{H \times W \times 3}$, through a well-defined projection function denoted as ${\xi}_{\mathcal{P}_{s}^{raw}, \mathcal{I}_s} \rightarrow \mathcal{D}_s$. In this particular scenario, the mapping function $\xi$ takes the form of a neural network commonly known as a depth network \cite{hu2021penet}.



\subsection{LiDAR Feature Extraction through Voxelization}

As part of the preprocessing phase, we transform the $s$-th input point cloud data ($\mathcal{P}_s^{raw}$) into a voxelized representation with dimensions $H_{v}\times W_{v} \times C_{v}$, denoted as $L_{s}$, and compute voxel features by averaging point-wise features for non-empty voxels \cite{shi2023pv}. To identify key points, we employ the furthest point sampling (FPS) method \cite{shi2023pv}, resulting in $\mathcal{K}$ key points ($L_s^{\mathcal{K}}$), where $\mathcal{K}$ equals 4096 for both experiments. Following that, we characterize non-empty voxels by computing the mean of characteristics such as 3D coordinates and reflectance values from all points within each voxel. The feature volumes of the point cloud then undergo transformation through a sequence of $3\times3\times3$ 3D sparse convolutions \cite{shi2023pv}, resulting in downsampled spatial resolutions of $1\times, 2\times, 4\times,$ and $8\times$. The sparse feature volumes may be conceptualized as ensembles of feature vectors that are linked to particular voxels. Ultimately, the final feature vectors for each voxel in the LiDAR sample are represented as $\mathbb{F}^{LiDAR} \in \mathbb{R}^{H\times W\times C}$.

\subsection{GCFAT}
\begin{figure}[h]
    \centering
    \scalebox{0.3}{
    \includegraphics{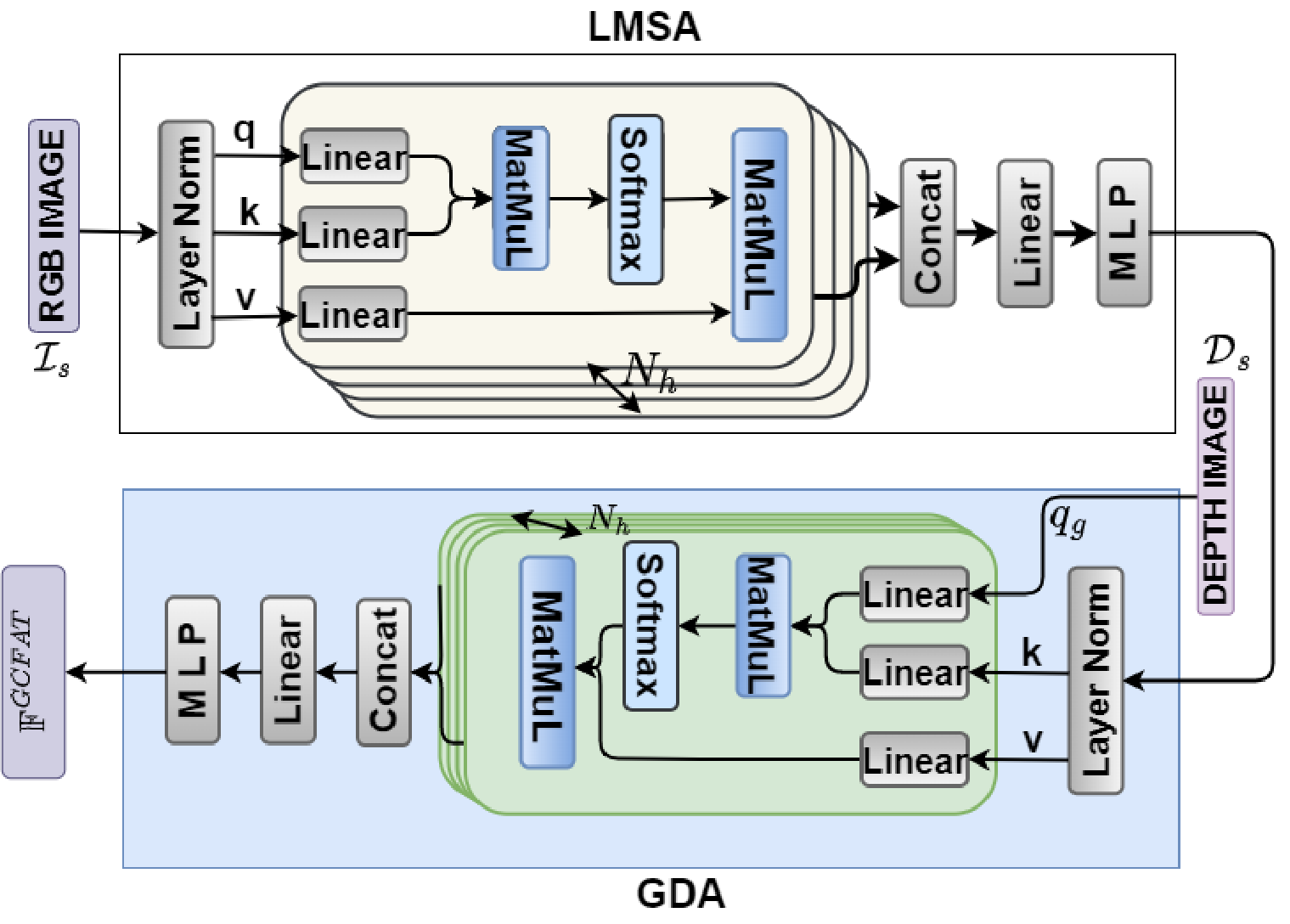}}
    \caption{GCFAT Comprises LMSA and GDA}
    \label{fig:GCFAT}
\end{figure}

We introduce a novel fusion alignment transformer called GCFAT, as depicted in Figure \ref{fig:GCFAT}, which integrates local features from the RGB image ($ x \in \mathcal{I}_s$) and global depth estimation ($x_d \in \mathcal{D}_s$) to produce aligned feature representations. Therefore,  each GCFAT stage comprises alternating local attention and global diffused attention (GDA) modules, which extract enriched feature representations. Local windows operate similarly to baseline vision transformers, such as the Swin Transformer \cite{liu2022swin}, Global Transformer \cite{hatamizadeh2022global} while the RGB images pass through a local query generator that utilizes the Local Multi-scale Attention (LMSA) module to extract local features and focus on different levels of detail \cite{liu2022swin}. 

\textbf{GDA} The LMSA mechanism is constrained to interrogating patches that fall within a designated local window. By contrast, the global attention mechanism possesses the capability to interrogate diverse modalities that are obtained from the computation of depth maps ($x_d$), while operating within a unified framework. Unlike other single modalities method, the computation of the global query element is pre-determined. Therefore, GDA uses global query tokens from depth maps ($x_d$) to interact with local key and value representations, while GCFAT efficiently captures local and spatial complexities by exchanging local and global self-attention blocks between the two modalities($\mathcal{I}_s,\mathcal{D}_s $). The GDA module enhances the global context by applying attention to the entire scene, considering the correlation between RGB-derived features (key-value pairs) and the initial depth map (query). Specifically, the global depth query, $q_g$, has dimensions ${B \times C \times h_p \times w_p}$, representing batch size ($B$), embedding patch dimension ($C$), and local patch window height and width ($h_p$ and $w_p$). To align with the total window count, $q_g$ is duplicated along the batch dimension, resulting in an augmented batch size of $B^\ast = B \times N^\ast$, where $N^\ast$ represents the number of local patch windows. In each local window, key and value are computed using a linear layer, efficiently extracting relevant information. Through the interaction of local windows and global depth query tokens, the GDA module extends its receptive field, attending to diverse regions in the input feature maps. Thus, the GDA module is expressed as,
\begin{align}\label{eq:GDA}
\text{GDA}(x, x_d) &= \mathcal{LN}(\alpha v) \nonumber \\
\text{where, } \alpha & = \text{Softmax}(g(q_g, k^T))
\end{align}
    
where, $q_g \in x_d$, $k \in x$, and $v \in x$ are the query, key and value for GDA respectively. The $g(\cdot)$ represents attentive function between two different modalities as described in Equation \ref{eq:GDA}. In addition, the 
 $\mathcal{LN}(\cdot)$ refers to different variants of normalization, we use the Layer Norm \cite{hatamizadeh2022global}. Consequently, we denote the extraction of image features based on the global depth query as $\mathbb{F}^{GCFAT} \in \mathbb{R}^{H\times W\times C}$.

\subsection{SFFA}
\begin{figure}[h]
    \centering
    \scalebox{1.5}{
    \includegraphics[width=0.25\textwidth]{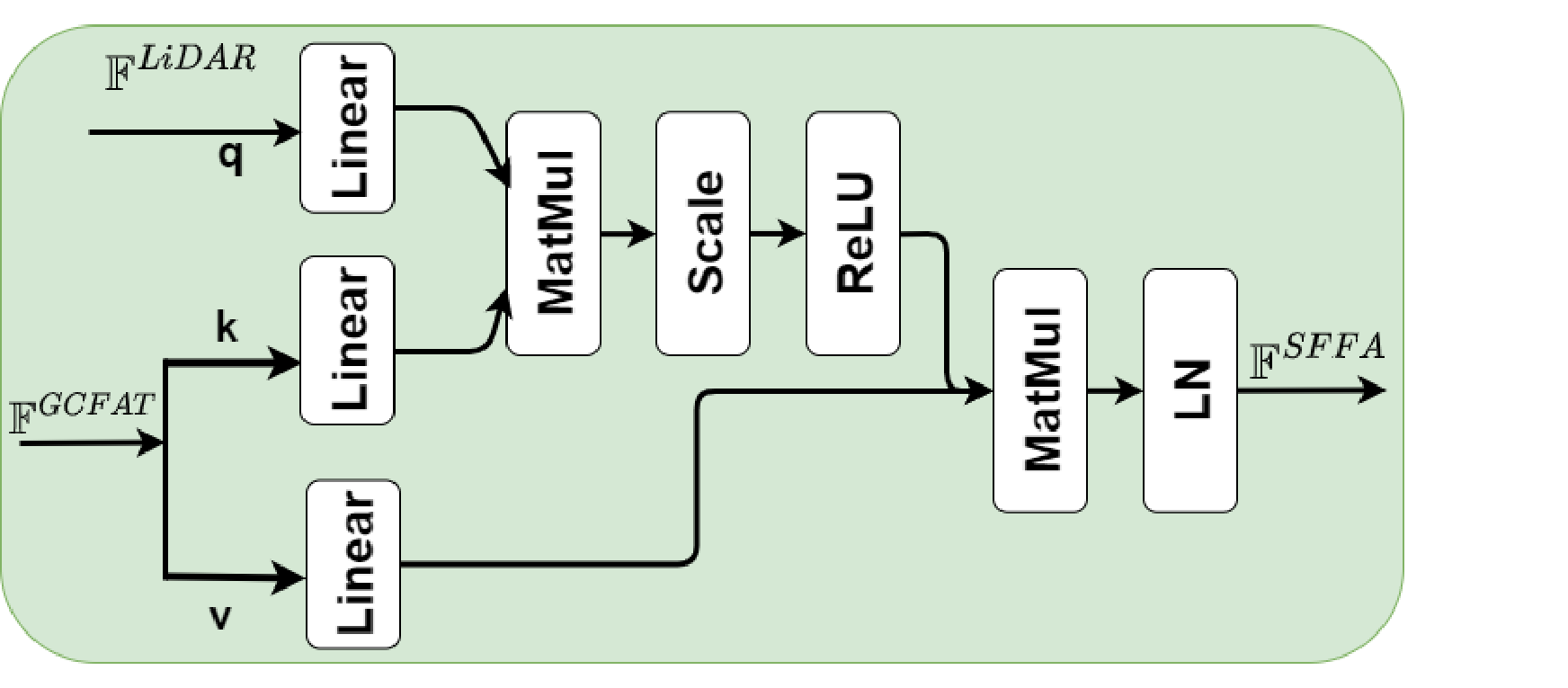}}
    \caption{SFFA combines extracted LiDAR and image features.}
    \label{fig:SFFA}
\end{figure}

Introducing a single-stage integration technique called SFFA as depicted in the Figure \ref{fig:SFFA}, which aligns sparse voxelized LiDAR features ($\mathbb{F}^{LiDAR}$) with image features from GCFAT ($\mathbb{F}^{GCFAT}$). Similar to GDA module, we treat the features extracted from GCFAT as keys($k$) to search for correspondences with queries($q$), thereby aligning LiDAR features ($\mathbb{F}^{LiDAR}$) with similar structures in $\mathbb{F}^{GCFAT}$ images.  To provide a clearer understanding of the key-query-value matching in our proposed SFFA mechanism, we describe it below,

\begin{align}\label{eq:SFFA_att}
\text{SFFA}(\mathbb{F}^{LiDAR}, \mathbb{F}^{GCFAT}) &= \mathcal{LN}(\beta v) \nonumber \\
\text{where, } \beta & = \text{ReLU}(f(q, k^T))
\end{align}
    
where, $q \in \mathbb{F}^{LiDAR}$, $k \in \mathbb{F}^{GCFAT}$, and $v \in \mathbb{F}^{GCFAT}$ are the query, key and value for SFFA respectively. The $f(\cdot)$ represents attentive function between two different modalities as described in Equation \ref{eq:SFFA_att}. Moreover, the ReLU block output and the original $\mathbb{F}^{GCFAT}$ values are merged through matrix multiplication, capturing correspondences and integrating 3D structure from $\mathbb{F}^{LiDAR}$ with texture and color details from $\mathbb{F}^{GCFAT}$. In addition, we follow the same normalisation in proposed AYDIV method $\mathcal{LN}(\cdot) = \text{RMSNorm}(\cdot)$ \cite{zhang2021sparse}. 
Hence, we represent the extraction of image features using the global LiDAR query as $\mathbb{F}^{SFFA} \in \mathbb{R}^{H\times W\times C}$.

\subsection{VGA}
\begin{figure}[h]
    \centering
    \scalebox{0.4}{
    \includegraphics{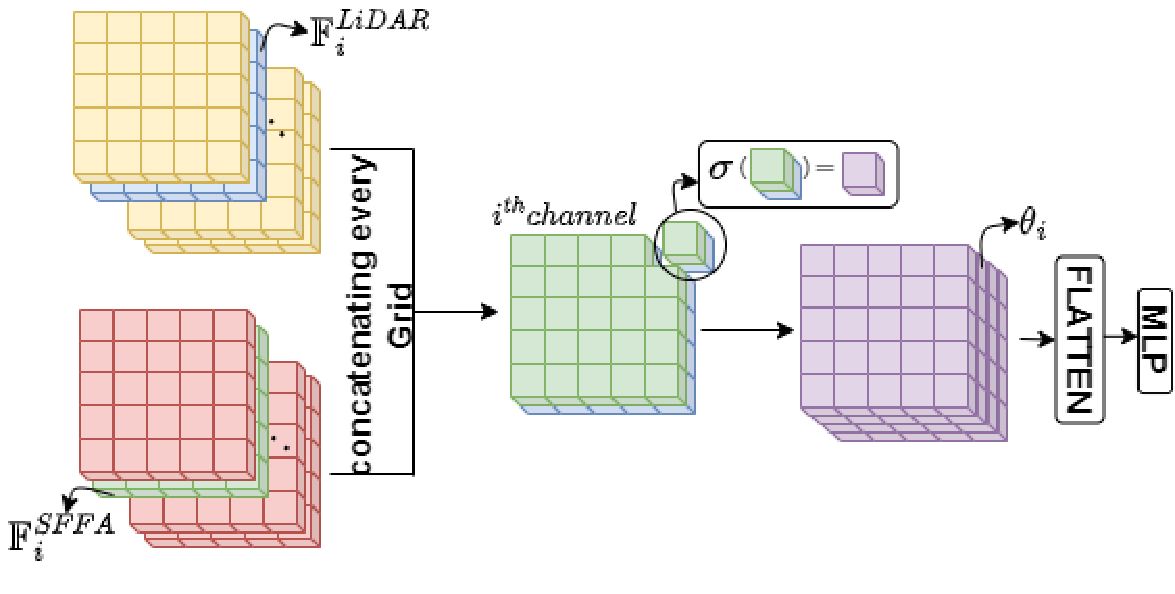}}
    \caption{VGA: Volumetric integration between $\mathbb{F}_i^{LiDAR}$ and $\mathbb{F}_i^{SFFA}$ though $i^{th}$ channel.}
    \label{fig:VGA}
\end{figure}
We present VGA in Figure \ref{fig:VGA}, a novel grid-wise fusion technique that integrates the along channel dimensionality between two modalities: pseudo Image RoI features (output from the SFFA module), denoted as $\mathbb{F}^{SFFA}$, and pseudo LiDAR RoI features, represented as $\mathbb{F}^{LiDAR}$. Therefore, we apply a fully MLP layer to generate a set of scalars  ($\theta_i^{LiDAR}$, $\theta_i^{SFFA}$) where both $\theta_i^{LiDAR}$ and $\theta_i^{SFFA}$ are learnable parameters. The fusion of ($\mathbb{F}^{LiDAR}$, $\mathbb{F}^{SFFA}$) is obtained by weighting them with ($\theta_i^{LiDAR}$, $\theta_i^{SFFA}$), resulting in the fused attentive grid feature $\mathbb{F}_{i} \in \mathbb{R}^{H\times W\times C}$. Mathematically, $\mathbb{F}_{i}$ is obtained as follows:

\begin{equation}
    (\theta^{LIDAR}_{i} , \theta^{SFFA}_{i})={\sigma}(MLP(CONCAT(\mathbb{F}^{LiDAR}_{i} , \mathbb{F}^{SFFA}_{i})))
\end{equation}

\begin{equation}
        \mathbb{F}_{i}=MLP(CONCAT(\theta^{LiDAR}_{i} \mathbb{F}^{LiDAR}_{i} , \theta^{SFFA}_{i} \mathbb{F}^{SFFA}_{i}))
\end{equation}

By transforming 2D images into 3D-like cloud structures, we can intricately combine the RoI features from both images and point clouds.

\subsection{Loss function} AYDIV uses Voxel R-CNN \cite{deng2021voxel} for RPN and RoI loss, in addition to using Fusion Loss \cite{zhang2021sparse} and transformer associated loss \cite{hatamizadeh2023global}.


\begin{table*}
\centering
\caption{Comparison of Model Performance for 3D Detection on the WOD Test Set. In the table, `L' and `I' denote LiDAR and camera sensors, respectively. `TTA' and `Ens' represent test-time augmentation and ensemble model outputs, indicated by \texttt{\#}}
\label{tab:AYDIV_test_set}
\scalebox{1}{
\begin{tabular}{l|l|l|ll|ll|ll}
\hline
\textbf{Method} & \textbf{Modality} & \textbf{ALL (mAPH)} & \multicolumn{2}{l|}{\textbf{VEH (AP/APH)}} & \multicolumn{2}{l|}{\textbf{PED (AP/APH)}} & \multicolumn{2}{l|}{\textbf{CYC (AP/APH)}} \\ \hline
 &  & L2 & \multicolumn{1}{l|}{L1} & L2 & \multicolumn{1}{l|}{L1} & L2 & \multicolumn{1}{l}{L1} & L2 \\ \hline

 AYDIV TTA \texttt{\#}  (Ours) & L+I & \textcolor{blue}{82.04} \textcolor{red}{(+1.02)}& \multicolumn{1}{l|}{\textcolor{blue}{89.12}/\textcolor{blue}{88.45}} & \textcolor{blue}{83.21}/\textcolor{blue}{82.03} & \multicolumn{1}{l|}{\textcolor{blue}{88.98}/\textcolor{blue}{87.01}} & \textcolor{blue}{85.63}/\textcolor{blue}{83.24} & \multicolumn{1}{l|}{\textcolor{blue}{84.35}/\textcolor{blue}{83.34}} & \textcolor{blue}{82.31}/\textcolor{blue}{80.87} \\ 

AYDIV (Ours) & L+I & 81.77 & \multicolumn{1}{l|}{88.69/88.02} & 83.01/81.98 & \multicolumn{1}{l|}{88.82/86.84} & 85.22/83.12 & \multicolumn{1}{l|}{83.74/83.21} & 81.84/80.20 \\  \hline
LoGoNet Ens\texttt{\#} \cite{li2023logonet}  & L+I & 81.02 & \multicolumn{1}{l|}{88.33/87.87} & 82.17/81.72 & \multicolumn{1}{l|}{\textcolor{blue}{88.98}/85.96} & 84.27/81.28 & \multicolumn{1}{l|}{83.10/82.16} & 80.93/80.06 \\ 
BEVFusion TTA\texttt{\#} \cite{liu2023bevfusion} & L+I & 79.97 & \multicolumn{1}{l|}{87.96/87.58} & 81.29/80.92 & \multicolumn{1}{l|}{87.64/85.04} & 82.19/79.65 & \multicolumn{1}{l|}{82.53/81.67} & 80.17/79.33 \\ 
LidarMultiNet TTA\texttt{\#} \cite{ye2023lidarmultinet} & L & 79.94 & \multicolumn{1}{l|}{87.64/87.26} & 80.73/80.36 & \multicolumn{1}{l|}{87.75/85.07} & 82.48/79.86 & \multicolumn{1}{l|}{82.77/81.84} & 80.50/79.59 \\ 
MPPNet Ens\texttt{\#} \cite{chen2022mppnet} & L & 79.60 & \multicolumn{1}{l|}{87.77/87.37} & 81.33/80.93 & \multicolumn{1}{l|}{87.92/85.15} & 82.86/80.14 & \multicolumn{1}{l|}{80.74/79.90} & 78.54/77.73 \\ 
MT-Net Ens\texttt{\#} \cite{chen2022mt} & L & 78.45 & \multicolumn{1}{l|}{87.11/86.69} & 80.52/80.11 & \multicolumn{1}{l|}{86.50/83.55} & 80.95/78.08 & \multicolumn{1}{l|}{80.50/79.43} & 78.22/77.17 \\ 
DeepFusion Ens\texttt{\#} \cite{li2022deepfusion} & L+I & 78.41 & \multicolumn{1}{l|}{86.45/86.09} & 79.43/79.09 & \multicolumn{1}{l|}{86.14/83.77} & 80.88/78.57 & \multicolumn{1}{l|}{80.53/79.80} & 78.29/77.58 \\ 
AFDetV2 Ens\texttt{\#} \cite{hu2022afdetv2} & L & 77.64 & \multicolumn{1}{l|}{85.80/85.41} & 78.71/78.34 & \multicolumn{1}{l|}{85.22/82.16} & 79.71/76.75 & \multicolumn{1}{l|}{81.20/80.30} & 78.70/77.83 \\ 
INT Ens\texttt{\#} \cite{xu2022int} & L & 77.21 & \multicolumn{1}{l|}{85.63/85.23} & 79.12/78.73 & \multicolumn{1}{l|}{84.97/81.87} & 79.35/76.36 & \multicolumn{1}{l|}{79.76/78.65} & 77.62/76.54 \\ 
HorizonLiDAR3D Ens\texttt{\#} \cite{ding20201st} & L+I & 77.11 & \multicolumn{1}{l|}{85.09/84.68} & 78.23/77.83 & \multicolumn{1}{l|}{85.03/82.10} & 79.32/76.50 & \multicolumn{1}{l|}{79.73/78.78} & 77.91/76.98 \\ \hline
LoGoNet \cite{li2023logonet}  & L+I & 77.10 & \multicolumn{1}{l|}{86.51/86.10} & 79.69/79.30 & \multicolumn{1}{l|}{86.84/84.15} & 81.55/78.91 & \multicolumn{1}{l|}{76.06/75.25} & 73.89/73.10 \\ 
BEVFusion \cite{liu2023bevfusion} & L+I & 76.33 & \multicolumn{1}{l|}{84.97/84.55} & 77.88/77.48 & \multicolumn{1}{l|}{84.72/81.97} & 79.06/76.41 & \multicolumn{1}{l|}{78.49/77.54} & 76.00/75.09 \\ 
CenterFormer \cite{zhou2022centerformer} & L & 76.29 & \multicolumn{1}{l|}{85.36/84.94} & 78.68/78.28 & \multicolumn{1}{l|}{85.22/82.48} & 80.09/77.42 & \multicolumn{1}{l|}{76.21/75.32} & 74.04/73.17 \\ 
MPPNet \cite{chen2022mppnet} & L & 75.67 & \multicolumn{1}{l|}{84.27/83.88} & 77.29/76.91 & \multicolumn{1}{l|}{84.12/81.52} & 78.44/75.93 & \multicolumn{1}{l|}{77.11/76.36} & 74.91/74.18 \\ 
DeepFusion \cite{li2022deepfusion} & L+I & 75.54 & \multicolumn{1}{l|}{83.25/82.82} & 76.11/75.69 & \multicolumn{1}{l|}{84.63/81.80} & 79.16/76.40 & \multicolumn{1}{l|}{77.81/76.82} & 75.47/74.51 \\ \hline
\end{tabular}
}
\label{tab:Waymo_test_set}
\end{table*}


\begin{table*}
\centering
\caption{Comparative Performance Analysis on the Waymo Validation Set for 3D Vehicle Detection (IoU = 0.7), Pedestrian Detection (IoU = 0.5), and Cyclist Detection (IoU = 0.5). PV-RCNN \cite{shi2020pv} is our baseline model. }
\scalebox{1.0}{
\begin{tabular}{l|l|l|ll|ll|ll}
\hline
\textbf{Method} & \textbf{Modality} & \textbf{ALL (mAPH)} & \multicolumn{2}{l|}{\textbf{VEH (AP/APH)}} & \multicolumn{2}{l|}{\textbf{PED (AP/APH)}} & \multicolumn{2}{l|}{\textbf{CYC (AP/APH)}} \\ \hline
 &  & L2 & \multicolumn{1}{l|}{L1} & L2 & \multicolumn{1}{l|}{L1} & L2 & \multicolumn{1}{l|}{L1} & L2 \\ \hline
SECOND \cite{yan2018second} & L & 57.23 & 72.27/71.69 & 63.85/63.33 & 68.70/58.18 & 60.72/51.31 & 60.62/59.28 & 58.34/57.05 \\ 
PointPillars \cite{lang2019pointpillars} & L & 57.53 & 71.60/71.00 & 63.10/62.50 & 70.60/56.70 & 62.90/50.20 & 64.40/62.30 & 61.90/59.90 \\ 
LiDAR-RCNN \cite{li2021lidar} & L & 60.10 & 73.50/73.00 & 64.70/64.20 & 71.20/58.70 & 63.10/51.70 & 68.60/66.90 & 66.10/64.40 \\ 

CenterPoint\cite{yin2021center} & L & 65.46 & - & -/66.20 & - & -/62.60 & - & -/67.60 \\ 
PointAugmenting \cite{wang2021pointaugmenting} & L+I & 66.70 & 67.4/- & 62.7/- & 75.04/- & 70.6/- & 76.29/- & 74.41/- \\ 
Pyramid-PV \cite{mao2021pyramid} & L & - & 76.30/75.68 & 67.23/66.68 & - & - & - & - \\ 
PDV \cite{hu2022point} & L & 64.25 & 76.85/76.33 & 69.30/68.81 & 74.19/65.96 & 65.85/58.28 & 68.71/67.55 & 66.49/65.36 \\ 
Graph-RCNN \cite{yang2022graph} & L & 70.91 & 80.77/80.28 & 72.55/72.10 & 82.35/76.64 & 74.44/69.02 & 75.28/74.21 & 72.52/71.49 \\ 
3D-MAN \cite{yang20213d} & L & - & 74.50/74.00 & 67.60/67.10 & 71.70/67.70 & 62.60/59.00 & - & - \\ 
Centerformer \cite{zhou2022centerformer} & L & 73.70 & 78.80/78.30 & 74.30/73.80 & 82.10/79.30 & 77.80/75.00 & 75.20/74.40 & 73.20/72.30 \\ 
DeepFusion \cite{li2022deepfusion} & L+I & - & 80.60/80.10 & 72.90/72.40 & 85.80/83.00 & 78.70/76.00 & - & - \\ 
MPPNet \cite{chen2022mppnet} & L & 74.22 & 81.54/81.06 & 74.07/73.61 & 84.56/81.94 & 77.20/74.67 & 77.15/76.50 & 75.01/74.38 \\ 
MPPNet \cite{chen2022mppnet} & L & 74.85 & 82.74/82.28 & 75.41/74.96 & 84.69/82.25 & 77.43/75.06 & 77.28/76.66 & 75.13/74.52 \\ 
LoGoNet \cite{li2023logonet} & L+I & 75.54 & 83.21/82.72 & 75.84/75.38 & 85.80/83.14 & 78.97/76.33 & 78.58/77.79 & 75.67/74.91 \\ \hline
Baseline\cite{shi2020pv} & L & 63.33 & 77.51/76.89 & 68.98/68.41 & 75.01/65.65 & 66.04/57.61 & 67.81/66.35 & 65.39/63.98 \\ 
AYDIV (ours)  & L+I & \textcolor{blue}{78.77} \textcolor{red}{(+15.44)} & \textcolor{blue}{86.36}/\textcolor{blue}{85.72} & \textcolor{blue}{81.04}/\textcolor{blue}{79.75} & \textcolor{blue}{88.20}/\textcolor{blue}{86.24} & \textcolor{blue}{82.79}/\textcolor{blue}{79.07} & \textcolor{blue}{81.73}/\textcolor{blue}{80.24} & \textcolor{blue}{78.32}/\textcolor{blue}{77.50} \\ \hline
\end{tabular}
}
\label{tab:Waymo_valid_set}
\end{table*}

\section{Experiments}
\subsection{Dataset details}

\textbf{WOD\cite{sun2020scalability}} leads in 3D object detection benchmarks, with 1,150 sequences, 200K+ frames, and a mix of LiDAR, images, and 3D bounding boxes. It consists of 798 training, 202 validation, and 150 testing sequences, with a 75-meter detection range and a 150m x 150m coverage area. We evaluate models using Average Precision (AP) and Average Precision weighted by Heading (APH)\cite{sun2020scalability, li2023logonet}. We present the results for both  \text{LEVEL\texttt{\string_}1 (L1)} and \text{LEVEL\texttt{\string_}2 (L2)} difficulty items, providing a comprehensive assessment and contrast of the models' performance. 

\textbf{AV2}\cite{wilson2023argoverse} validates our long-range experiments, emphasizing a 200-meter perception range and 400m $\times$ 400m coverage area. It includes 1,000 sequences: 700 for training, 150 for validation, and 150 for testing\cite{Argoverse2}. AV2 encompasses 30 object classes, but we evaluate using the 20 head classes, excluding the 10 tail classes, with the Average Precision (AP) metric\cite{yin2021center, fan2022fully, li2022deepfusion}.



\subsection{Implementation Details}

\textbf{Network Architecture.} The LiDAR block in AYDIV is based on the Voxel-RCNN architecture\cite{shi2020pv}. We calculate depth information using both modalities, following the depth network from \cite{hu2021penet}. We employ augmentation methods such as rotation, flipping, global scaling, local noise, and training with similar classes\cite{shi2020pv}. The Voxelization features are extracted using a 3D voxel CNN with four levels, featuring dimensions of 16, 32, 64, and 64\cite{shi2020pv}. In the GDA module, the number of attention  8 heads ($N_h$) with projection dimension of $C = 64$,  and patch sizes of attention windows ($h_p$ and $w_p$) set to 7. We apply 30\% dropout to the attention affinity matrix for regularization. The MLP layer after the GDA module and the SFFA shares the same structure, with one head in SFFA. RoI-grid pooling uses VGA, selecting $6\times6\times6$ grid points within each 3D proposal, followed by an MLP layer with 64 filters without dropout.



\textbf{Training and Inference Details.}
AYDIV is trained from scratch using the ADAM optimizer with a batch size of 32 and a learning rate of 0.01 for 100 epochs. In the proposal refinement stage, 128 proposals are randomly selected, maintaining a 1:1 ratio between positive and negative proposals. We set the voxel size to (0.1m, 0.1m, 0.15m) for both datasets to enhance spatial resolution\cite{li2023logonet, li2022deepfusion,liu2023flatformer}. During inference, NMS is applied twice: first with an IoU threshold of 0.7 to choose the top-100 region proposals as inputs for the detection head, and then, after refinement, with an IoU threshold of 0.1 to remove redundant predictions\cite{li2022deepfusion}.

\begin{figure}
    \centering
    \subfloat[ AYDIV]{%
        \includegraphics[width=0.23\textwidth, height=3.2cm]{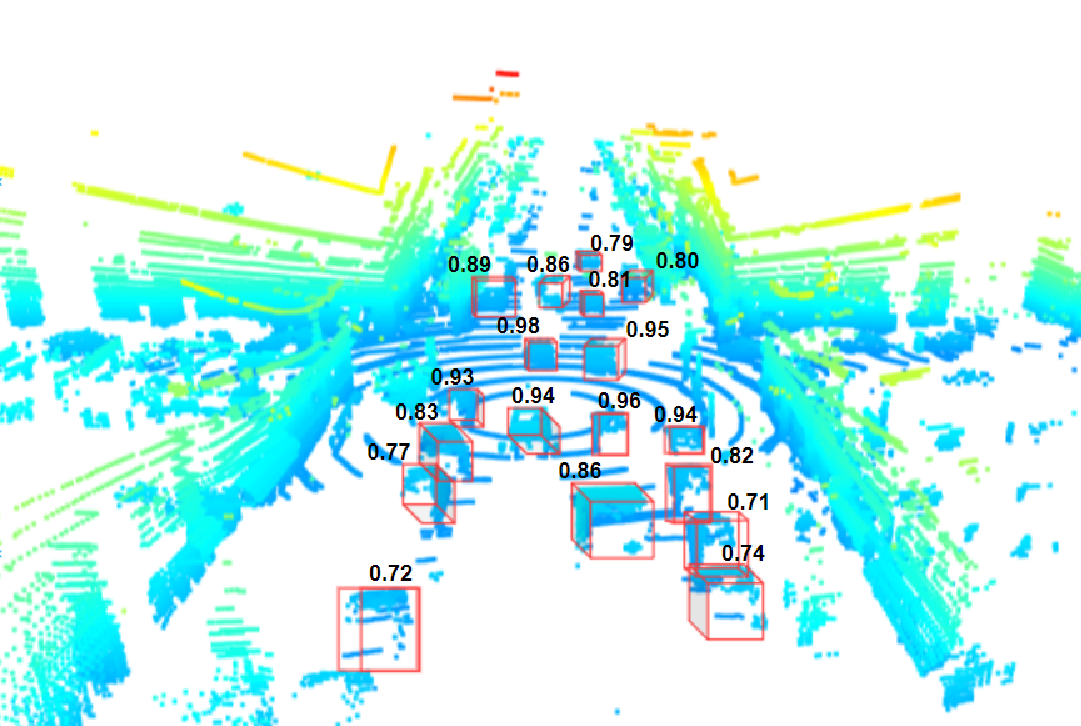}
        \includegraphics[width=0.23\textwidth, height=2.9cm]{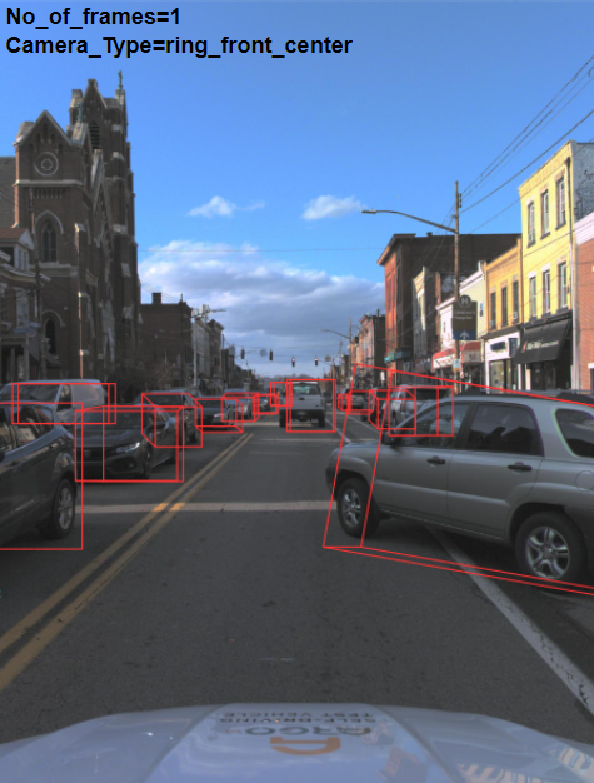}
    }
    \hspace{1em} 
    \subfloat[ LoGoNet]{%
        \includegraphics[width=0.23\textwidth, height=3.2cm]{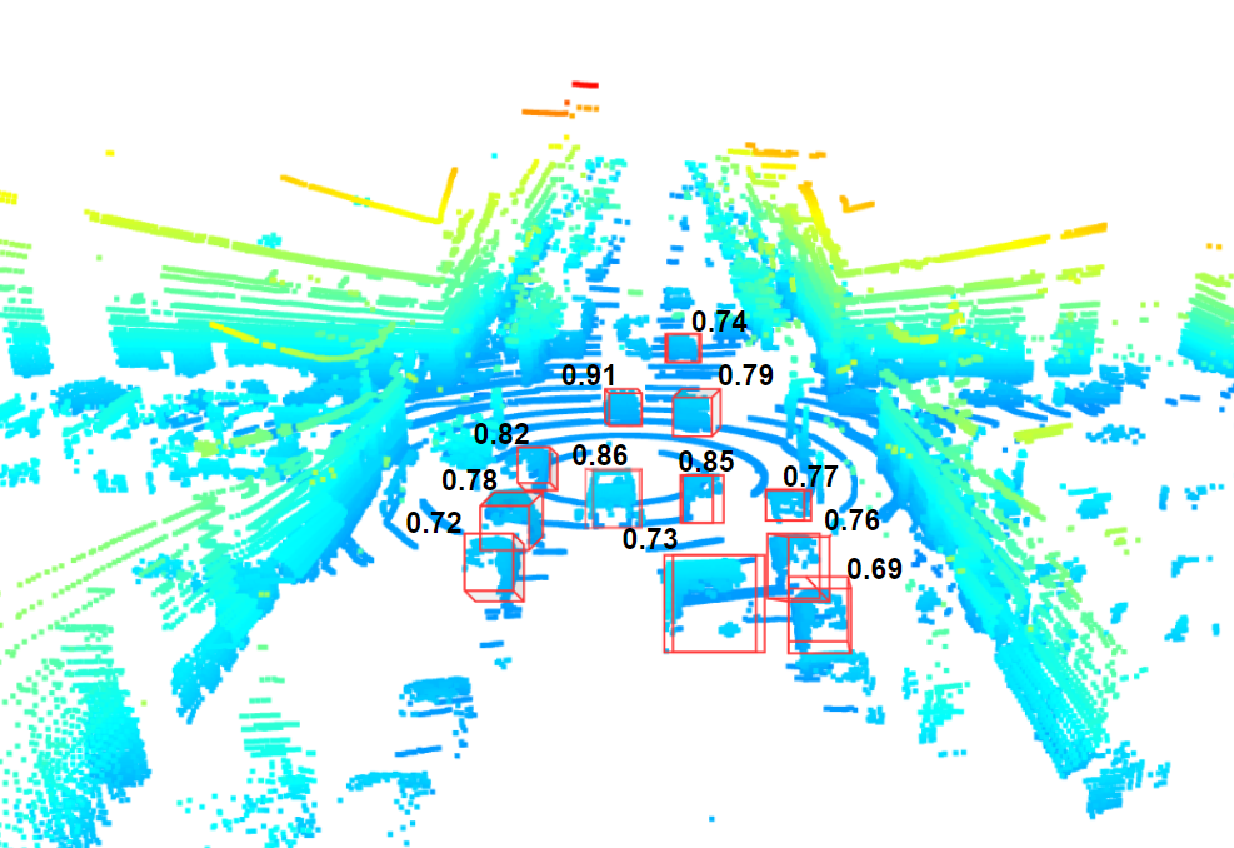}
        \includegraphics[width=0.23\textwidth, height=2.9cm]{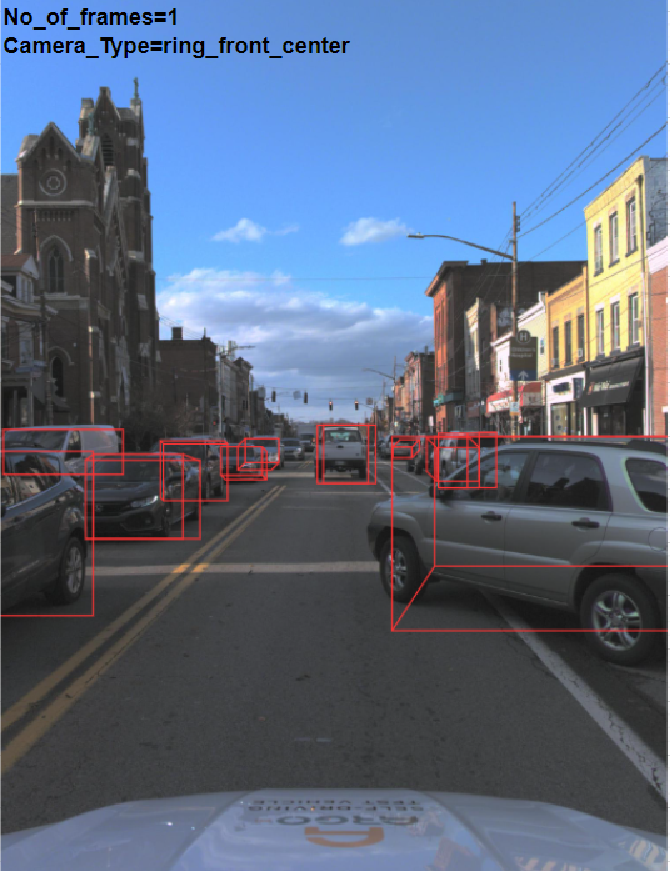}
    }
    \caption{A qualitative comparison of long-range 3D object detection. BEV maps on left, 2D image from camera 1 on right. Red: Predicted boxes. Black: Detection score.}
    \label{fig:compare_av2}
\end{figure}



\begin{table*}
\centering
\caption{The table presents AV2 validation split performance across categories. Significant improvements (in bold) are seen in some categories. The first three models use `L' and the remaining three use `L+I'. \texttt{\#}: Simulated on the same enviorments}

\scalebox{0.8}{
\begin{tabular}{l|cccccccccccccccccccc|c}
\hline
Methods   &\begin{sideways}\textbf{Vehicle}\end{sideways} & \begin{sideways}\textbf{Bus} \end{sideways}  & \begin{sideways}\textbf{Pedestrian}\end{sideways} & \begin{sideways}\textbf{Stop Sign}\end{sideways} & \begin{sideways}Box Truck\end{sideways} & \begin{sideways}\textbf{Bollard}\end{sideways} & \begin{sideways}\textbf{C-Barrel}\end{sideways} & \begin{sideways}Motorcyclist\end{sideways} & \begin{sideways}\textbf{MPC-Sign}\end{sideways} & \begin{sideways}\textbf{Motorcycle} \end{sideways} & \begin{sideways}\textbf{Bicycle} \end{sideways} & \begin{sideways}\textbf{A-Bus} \end{sideways}& \begin{sideways}\textbf{School Bus} \end{sideways} & \begin{sideways}\textbf{Truck Cab} \end{sideways} & \begin{sideways}\textbf{C-Cone} \end{sideways}& \begin{sideways}\textbf{V-Trailer} \end{sideways} & \begin{sideways}\textbf{Sign} \end{sideways} & \begin{sideways}\textbf{Large Vehicle} \end{sideways} & \begin{sideways}\textbf{Stroller} \end{sideways} & \begin{sideways}\textbf{Bicyclist} \end{sideways} & \begin{sideways}\textbf{AP}\end{sideways}\\ 
\hline
\textit{Precision} &         &         &       &            &           &           &         &          &              &          &            &         &       &            &           &        &           &      &               &          &           \\ \hline
CenterPoint \cite{yin2021center}          & 61.0    & 36.0 & 33.0       & 28.0      & 26.0      & 25.0    & 22.5     & 16.0         & 16.0     & 12.5       & 9.5     & 8.5   & 7.5        & 8.0       & 8.0    & 7.0       & 6.5  & 3.0           & 2.0      & 14       & 17.5
    \\

CenterPoint\texttt{+} \cite{fan2022fully}          & 67.6 & 38.9 & 46.5 & 16.9 & 37.4 & 40.1 & 32.2 & 28.6 & 27.4 & 33.4 & 24.5 & 8.7 & 25.8 & \textcolor{blue}{22.6} & 29.5 & 22.4 & 6.3 & 3.9 & 0.5 & 20.1  & 26.67    \\

FSD \cite{fan2022fully}                 & 67.1 & 39.8 & 57.4 & 21.3 & 38.3 & 38.3 & 38.1 & 30.0 & 23.6 & 38.1 & 25.5 & 15.6 & 30.0 & 20.1 & 38.9 & 23.9 & 7.9 & 5.1 & 5.7 & 27.0 & 29.58   \\ \hline


BEVFusion \cite{liu2023bevfusion} \texttt{\#}        & 67.2   &   39.8   &   58.1   &   31.9   &   36.3   &   35.2   &   36.7   &   34.1   &   26.1   &   46.8   &   33.6   &   21.2   &   22.2   &   16.9   &   31.2   &   22.8   &   13.2   &   5.4   &   9.6   &   32.6 & 31.05 \\ 

DeepFusion \cite{li2022deepfusion} \texttt{\#}         & 70.7 & 42.3 & 62.1 & 32.8 & 40.8 & 40.0 & 42.2 & \textcolor{blue}{42.6} & 28.3 & 50.1 & 40.1 & 21.7 & 29.7 & 17.6 & 40.2 & 25.3 & 14.7 & 7.9 & 10.7 & 35.1 & 34.74     \\

LoGoNet \cite{li2023logonet} \texttt{\#}          & 73.6 & 42.5 & 64.4 & 32.9 & \textcolor{blue}{41.5} & 40.1 & 42.6 & 42.3 & 28.8 & 49.3 & \textcolor{blue}{41.3} & 20.7 & 25.9 & 18.5 & 39.8 & 25.2 & 15.5 & 7.2 & \textcolor{blue}{10.8} & 35.4  &  34.91  \\ \hline


AYDIV (ours)       &  \textcolor{blue}{76.2}    & \textcolor{blue}{47.1}  & \textcolor{blue}{67.4}      & \textcolor{blue}{38.6}      & 39.1     & \textcolor{blue}{42.3}    & \textcolor{blue}{44.8}     & 41.0         & \textcolor{blue}{30.6}    & \textcolor{blue}{52.7}       & 40.6    & \textcolor{blue}{25.4}  & \textcolor{blue}{30.9}       & 21.8      & \textcolor{blue}{41.4}   & \textcolor{blue}{30.7}      & \textcolor{blue}{19.8}    & \textcolor{blue}{10.9 }         & \textcolor{blue}{10.8}     & \textcolor{blue}{39.8 }     & \textcolor{blue}{37.70}\textcolor{red}{(+2.79)} \\ \hline
\end{tabular}}
\label{tab:AV2-C performance}
\end{table*}

\subsection{Performance on WOD and AV2}

\textbf{WOD.} The performance of AYDIV on the WOD test and validation set is detailed in Table \ref{tab:Waymo_test_set} and Table \ref{tab:Waymo_valid_set}, respectively. As illustrated in Table \ref{tab:Waymo_test_set}, AYDIV stands out, achieving superior performance over other leading methods for both L1 and L2 difficulties. When compared to the LoGoNet\cite{li2023logonet}, AYDIV showcases significant improvements, all achieved without the use of ensemble techniques or Test Time Augmentation (TTA).  More specifically, our non-TTA version of AYDIV outperforms the LoGoNet\cite{li2023logonet} by margins of 2.18 AP/L1, 1.92 APH/L1, 3.21 AP/L2, and 2.68 APH/L2 for vehicles; 1.98 AP/L1, 2.69 APH/L1, 3.67 AP/L2, and 4.21 APH/L2 for pedestrians; and 7.68 AP/L1, 7.96 APH/L1, 7.95 AP/L2, and 7.10 APH/L2 for cyclists, culminating in an aggregate enhancement of 4.67 mAPH/L2. Using the TTA version, AYDIV outperforms LoGoNet-Ens\cite{li2023logonet} by 0.79 AP/L1, 0.58 APH/L1, 1.04 AP/L2, and 0.31 APH/L2 in the vehicle class; 0.05 AP/L1, 1.05 APH/L1, 1.36 AP/L2, and 1.96 APH/L2 in the pedestrian category; and 1.25 AP/L1, 1.18 APH/L1, 1.38 AP/L2, and 0.81 APH/L2 for cyclists, resulting in an overall growth of 1.24\% mAPH/L2.

Table \ref{tab:Waymo_valid_set} provides a comprehensive comparison of model performance for 3D detection on the WOD validation set. Notably, AYDIV demonstrates significant improvements across different difficulty levels. In the L1 difficulty level, it outperforms LoGoNet\cite{li2023logonet} validation results on WOD by margins of 3.15 AP/L1, 3.00 APH/L1, 5.20 AP/L2, and 4.37 APH/L2 for vehicles; 2.40 AP/L1, 3.10 APH/L1, 3.82 AP/L2, and 2.74 APH/L2 for pedestrians; and 3.15 AP/L1, 2.45 APH/L1, 2.65 AP/L2, and 2.59 APH/L2 for cyclists, leading in an aggregate enhancement of 4.27\% mAPH/L2. These enhancements demonstrate AYDIV's ability in accurately identifying all classes, highlighting the potential of multi-modal feature alignment in refining 3D object detection.

\textbf{AV2.} Table \ref{tab:AV2-C performance} represents the performance of AYDIV with other state-of-the-art methods where we have considered both modalities performance. With the introduction of the modified version, CenterPoint$\textit{+}$, we observed a remarkable 52.4\% improvement in AP compared to its previous version,  CenterPoint. Additionally, the FSD method showed a significant performance jump in single modality-based 3D OD. The FSD method demonstrated an improved performance by 10.91\% in AP compared to CenterPoint$\textit{+}$. BEVFusion enhanced AP by 4.73\% relative to FSD when both modalities were considered. When cross-former-based feature fusion was added to DeepFusion via feature alignment, AP increased by 10.62\% compared to BEVFusion. While considering local-global attention mechanisms in LoGoNet, a negligible performance enhancement of 0.5\% in AP value was observed. Our proposed AYDIV, which incorporates three attention components, outperformed LoGoNet with an AP of 37.70, a 7.40\% improvement. For a better understanding of AYDIV's performance, we have compared performance in Figure \ref{fig:compare_av2}, where it is evident that AYDIV can detect more objects with a high confidence score in BEV maps.

\section{Ablation Studies on WOD}

\textbf{(A) Influence of each component.} Table \ref{tab:each_component} summarizes the impact of individual components on AYDIV model performance in two scenarios. Without SFFA, using only GCFAT and VGA results in a performance drop of 4.51\% for vehicles, 7.78\% for pedestrians, and 5.64\% for cyclists compared to using all components. This is due to limitations in the sparse LiDAR feature extractor ($\mathbb{F}^{LiDAR}$), which fails to achieve optimal fusion alignment despite estimating depth ($\mathcal{D}_{s}$) using LiDAR on images. When GCFAT is excluded, the performance drop is more substantial: 9.48\% for vehicles, 23.97\% for pedestrians, and 14.18\% for cyclists, despite projecting the image feature to match LiDAR feature, causing it to behave like a conventional voxelized LiDAR-based detector.

\begin{table}[h]
\centering
\caption{Influence of each component in AYDIV WOD testing set in L2 difficulty}
\begin{tabular}{lll|lll} \\ \hline
\multicolumn{3}{l}{Components}                                                             & \multicolumn{3}{l}{APH (L2)} \\ \hline
GCFAT                     & SFFA                               & VGA                       & VEH      & PED     & CYC     \\ \hline
\checkmark &                                    & \checkmark & 77.47    & 76.20   & 74.56   \\
                          & \checkmark          & \checkmark & 72.05    & 60.01   & 66.02   \\
\checkmark & \checkmark & \checkmark & \textcolor{blue}{81.98}    & \textcolor{blue}{83.98}   & \textcolor{blue}{80.20}    \\ \hline
\end{tabular}
\label{tab:each_component}

\end{table}

\begin{table}[h]
\centering
\caption{Different Vision Transformer performance on the Waymo Testing Set in L2 difficulty while excluding $\mathcal{D}_s$}
\begin{tabular}{lll|lll} \\ \hline
\multicolumn{3}{l|}{Vision Transformer}                              & \multicolumn{3}{l|}{APH (L2)} \\ \hline
\multicolumn{3}{l|}{}                                                & VEH      & PED     & CYC     \\ \hline
\multicolumn{3}{l|}{SwinV2 $\setminus \mathcal{D}_s$+SFFA+VGA} & 76.21    & 78.45   & 76.23   \\
\multicolumn{3}{l|}{GCVIT $\setminus \mathcal{D}_s$+SFFA+VGA}  & 77.46    & 78.93   & 77.19   \\ \hline
\multicolumn{3}{l|}{GCFAT+SFFA+VGA}                                  & \textcolor{blue}{81.98}    & \textcolor{blue}{83.98}   & \textcolor{blue}{80.20}  \\ \hline
\end{tabular}
\label{tab:vision_performance}
\end{table}

\textbf{(B) Importance of $\mathcal{D}_s$ in AYDIV.}
We evaluated two popular vision transformers models, SwinV2 \cite{liu2022swin} and GCVIT \cite{hatamizadeh2022global}, in conjunction with the SFFA and VGA components, while excluding depth information ($\mathcal{D}_s$). The results, in Table~\ref{tab:vision_performance}, showed that using SwinV2 reduced our model's performance by 7.04\% for vehicles, 6.58\% for pedestrians, and 4.95\% for cyclists compared to AYDIV. When we used GCVIT with fused conv2D, the performance improved compared to SwinV2 but still didn't reach AYDIV's level, with drops of 5.51\% for vehicles, 6.01\% for pedestrians, and 3.75\% for cyclists. Despite including other alignment methods, the absence of global context LiDAR information (disparity with images) leads to a noticeable decline in detection performance. This highlights the importance of using $\mathcal{D}_s$ as a global query to minimize disparity and enhance performance.

\section{CONCLUSIONS}
We introduced AYDIV, a 3D multi-modal object detection method based on transformers, consisting of three key components: GCFAT, SFFA, and VGA. These components were designed to capture both local and global dependencies, thereby enhancing the efficacy of 3D detection at both short and long distances. To determine the efficacy of AYDIV, we conducted comprehensive experiments on the WOD and AV2 benchmark datasets. AYDIV demonstrated its efficacy in multi-modal object detection by achieving competitive performance when compared to state-of-the-art methods. In addition, we conducted comprehensive ablation experiments to compare the effect of each proposed component on AYDIV's performance to other transformer-based techniques. 

In the future, AYDIV could be expanded to include robustness analysis, where we would examine both natural robustness and adversarial robustness conditions. While we have applied it in the context of autonomous vehicle data, the fusion alignment method can potentially find applications in other safety-critical domains, such as autonomous airports, where air traffic controllers depend on precise 3D detection methods to make critical decisions for complex tasks.





\bibliographystyle{IEEEtran}
\bibliography{icra}

\begin{thebibliography}{10}
\providecommand{\url}[1]{#1}
\csname url@samestyle\endcsname
\providecommand{\newblock}{\relax}
\providecommand{\bibinfo}[2]{#2}
\providecommand{\BIBentrySTDinterwordspacing}{\spaceskip=0pt\relax}
\providecommand{\BIBentryALTinterwordstretchfactor}{4}
\providecommand{\BIBentryALTinterwordspacing}{\spaceskip=\fontdimen2\font plus
\BIBentryALTinterwordstretchfactor\fontdimen3\font minus \fontdimen4\font\relax}
\providecommand{\BIBforeignlanguage}[2]{{%
\expandafter\ifx\csname l@#1\endcsname\relax
\typeout{** WARNING: IEEEtran.bst: No hyphenation pattern has been}%
\typeout{** loaded for the language `#1'. Using the pattern for}%
\typeout{** the default language instead.}%
\else
\language=\csname l@#1\endcsname
\fi
#2}}
\providecommand{\BIBdecl}{\relax}
\BIBdecl

\bibitem{arnold2019survey}
E.~Arnold, O.~Y. Al-Jarrah, M.~Dianati, S.~Fallah, D.~Oxtoby, and A.~Mouzakitis, ``A survey on 3d object detection methods for autonomous driving applications,'' \emph{IEEE Transactions on Intelligent Transportation Systems}, vol.~20, no.~10, pp. 3782--3795, 2019.

\bibitem{qi2017pointnet}
C.~R. Qi, H.~Su, K.~Mo, and L.~J. Guibas, ``Pointnet: Deep learning on point sets for 3d classification and segmentation,'' in \emph{Proceedings of the IEEE conference on computer vision and pattern recognition}, 2017, pp. 652--660.

\bibitem{qi2017pointnet++}
C.~R. Qi, L.~Yi, H.~Su, and L.~J. Guibas, ``Pointnet++: Deep hierarchical feature learning on point sets in a metric space,'' \emph{Advances in neural information processing systems}, vol.~30, 2017.

\bibitem{lang2019pointpillars}
A.~H. Lang, S.~Vora, H.~Caesar, L.~Zhou, J.~Yang, and O.~Beijbom, ``Pointpillars: Fast encoders for object detection from point clouds,'' in \emph{Proceedings of the IEEE/CVF conference on computer vision and pattern recognition}, 2019, pp. 12\,697--12\,705.

\bibitem{wang2021pointaugmenting}
C.~Wang, C.~Ma, M.~Zhu, and X.~Yang, ``Pointaugmenting: Cross-modal augmentation for 3d object detection,'' in \emph{Proceedings of the IEEE/CVF Conference on Computer Vision and Pattern Recognition}, 2021, pp. 11\,794--11\,803.

\bibitem{li2022deepfusion}
Y.~Li, A.~W. Yu, T.~Meng, B.~Caine, J.~Ngiam, D.~Peng, J.~Shen, B.~Wu, Y.~Lu, D.~Zhou, Q.~V. Le, A.~Yuille, and M.~Tan, ``Deepfusion: Lidar-camera deep fusion for multi-modal 3d object detection,'' 2022.

\bibitem{vora2020pointpainting}
S.~Vora, A.~H. Lang, B.~Helou, and O.~Beijbom, ``Pointpainting: Sequential fusion for 3d object detection,'' in \emph{Proceedings of the IEEE/CVF conference on computer vision and pattern recognition}, 2020, pp. 4604--4612.

\bibitem{9578812}
C.~Wang, C.~Ma, M.~Zhu, and X.~Yang, ``Pointaugmenting: Cross-modal augmentation for 3d object detection,'' in \emph{2021 IEEE/CVF Conference on Computer Vision and Pattern Recognition (CVPR)}, 2021, pp. 11\,789--11\,798.

\bibitem{huang2020epnet}
T.~Huang, Z.~Liu, X.~Chen, and X.~Bai, ``Epnet: Enhancing point features with image semantics for 3d object detection,'' 2020.

\bibitem{liang2020deep}
M.~Liang, B.~Yang, S.~Wang, and R.~Urtasun, ``Deep continuous fusion for multi-sensor 3d object detection,'' 2020.

\bibitem{chen2017multiview}
X.~Chen, H.~Ma, J.~Wan, B.~Li, and T.~Xia, ``Multi-view 3d object detection network for autonomous driving,'' 2017.

\bibitem{ku2018joint}
J.~Ku, M.~Mozifian, J.~Lee, A.~Harakeh, and S.~L. Waslander, ``Joint 3d proposal generation and object detection from view aggregation,'' in \emph{2018 IEEE/RSJ International Conference on Intelligent Robots and Systems (IROS)}.\hskip 1em plus 0.5em minus 0.4em\relax IEEE, 2018, pp. 1--8.

\bibitem{liang2020multitask}
M.~Liang, B.~Yang, Y.~Chen, R.~Hu, and R.~Urtasun, ``Multi-task multi-sensor fusion for 3d object detection,'' 2020.

\bibitem{deng2021voxel}
J.~Deng, S.~Shi, P.~Li, W.~Zhou, Y.~Zhang, and H.~Li, ``Voxel r-cnn: Towards high performance voxel-based 3d object detection,'' 2021.

\bibitem{chen2022mppnet}
X.~Chen, S.~Shi, B.~Zhu, K.~C. Cheung, H.~Xu, and H.~Li, ``Mppnet: Multi-frame feature intertwining with proxy points for 3d temporal object detection,'' in \emph{European Conference on Computer Vision}.\hskip 1em plus 0.5em minus 0.4em\relax Springer, 2022, pp. 680--697.

\bibitem{hu2022afdetv2}
Y.~Hu, Z.~Ding, R.~Ge, W.~Shao, L.~Huang, K.~Li, and Q.~Liu, ``Afdetv2: Rethinking the necessity of the second stage for object detection from point clouds,'' in \emph{Proceedings of the AAAI Conference on Artificial Intelligence}, vol.~36, no.~1, 2022, pp. 969--979.

\bibitem{shi2020point}
W.~Shi and R.~Rajkumar, ``Point-gnn: Graph neural network for 3d object detection in a point cloud,'' in \emph{Proceedings of the IEEE/CVF conference on computer vision and pattern recognition}, 2020, pp. 1711--1719.

\bibitem{fan2022embracing}
L.~Fan, Z.~Pang, T.~Zhang, Y.-X. Wang, H.~Zhao, F.~Wang, N.~Wang, and Z.~Zhang, ``Embracing single stride 3d object detector with sparse transformer,'' in \emph{Proceedings of the IEEE/CVF Conference on Computer Vision and Pattern Recognition}, 2022, pp. 8458--8468.

\bibitem{yin2021center}
T.~Yin, X.~Zhou, and P.~Krahenbuhl, ``Center-based 3d object detection and tracking,'' in \emph{Proceedings of the IEEE/CVF conference on computer vision and pattern recognition}, 2021, pp. 11\,784--11\,793.

\bibitem{bewley2020range}
A.~Bewley, P.~Sun, T.~Mensink, D.~Anguelov, and C.~Sminchisescu, ``Range conditioned dilated convolutions for scale invariant 3d object detection,'' \emph{arXiv preprint arXiv:2005.09927}, 2020.

\bibitem{chen2016monocular}
X.~Chen, K.~Kundu, Z.~Zhang, H.~Ma, S.~Fidler, and R.~Urtasun, ``Monocular 3d object detection for autonomous driving,'' in \emph{Proceedings of the IEEE conference on computer vision and pattern recognition}, 2016, pp. 2147--2156.

\bibitem{tao2023pseudo}
C.~Tao, J.~Cao, C.~Wang, Z.~Zhang, and Z.~Gao, ``Pseudo-mono for monocular 3d object detection in autonomous driving,'' \emph{IEEE Transactions on Circuits and Systems for Video Technology}, 2023.

\bibitem{ku2019monocular}
J.~Ku, A.~D. Pon, and S.~L. Waslander, ``Monocular 3d object detection leveraging accurate proposals and shape reconstruction,'' in \emph{Proceedings of the IEEE/CVF conference on computer vision and pattern recognition}, 2019, pp. 11\,867--11\,876.

\bibitem{shi2023multivariate}
X.~Shi, Z.~Chen, and T.-K. Kim, ``Multivariate probabilistic monocular 3d object detection,'' in \emph{Proceedings of the IEEE/CVF Winter Conference on Applications of Computer Vision}, 2023, pp. 4281--4290.

\bibitem{tao2023weakly}
R.~Tao, W.~Han, Z.~Qiu, C.-z. Xu, and J.~Shen, ``Weakly supervised monocular 3d object detection using multi-view projection and direction consistency,'' in \emph{Proceedings of the IEEE/CVF Conference on Computer Vision and Pattern Recognition}, 2023, pp. 17\,482--17\,492.

\bibitem{qi2018frustum}
C.~R. Qi, W.~Liu, C.~Wu, H.~Su, and L.~J. Guibas, ``Frustum pointnets for 3d object detection from rgb-d data,'' in \emph{Proceedings of the IEEE conference on computer vision and pattern recognition}, 2018, pp. 918--927.

\bibitem{wang2019frustum}
Z.~Wang and K.~Jia, ``Frustum convnet: Sliding frustums to aggregate local point-wise features for amodal 3d object detection,'' in \emph{2019 IEEE/RSJ International Conference on Intelligent Robots and Systems (IROS)}.\hskip 1em plus 0.5em minus 0.4em\relax IEEE, 2019, pp. 1742--1749.

\bibitem{liu2023flatformer}
Z.~Liu, X.~Yang, H.~Tang, S.~Yang, and S.~Han, ``Flatformer: Flattened window attention for efficient point cloud transformer,'' in \emph{Proceedings of the IEEE/CVF Conference on Computer Vision and Pattern Recognition}, 2023, pp. 1200--1211.

\bibitem{liang2018deep}
M.~Liang, B.~Yang, S.~Wang, and R.~Urtasun, ``Deep continuous fusion for multi-sensor 3d object detection,'' in \emph{Proceedings of the European conference on computer vision (ECCV)}, 2018, pp. 641--656.

\bibitem{piergiovanni20214d}
A.~Piergiovanni, V.~Casser, M.~S. Ryoo, and A.~Angelova, ``4d-net for learned multi-modal alignment,'' in \emph{Proceedings of the IEEE/CVF International Conference on Computer Vision}, 2021, pp. 15\,435--15\,445.

\bibitem{zheng2022beyond}
C.~Zheng, X.~Yan, H.~Zhang, B.~Wang, S.~Cheng, S.~Cui, and Z.~Li, ``Beyond 3d siamese tracking: A motion-centric paradigm for 3d single object tracking in point clouds,'' in \emph{Proceedings of the IEEE/CVF Conference on Computer Vision and Pattern Recognition}, 2022, pp. 8111--8120.

\bibitem{li2023logonet}
X.~Li, T.~Ma, Y.~Hou, B.~Shi, Y.~Yang, Y.~Liu, X.~Wu, Q.~Chen, Y.~Li, Y.~Qiao \emph{et~al.}, ``Logonet: Towards accurate 3d object detection with local-to-global cross-modal fusion,'' in \emph{Proceedings of the IEEE/CVF Conference on Computer Vision and Pattern Recognition}, 2023, pp. 17\,524--17\,534.

\bibitem{hu2021penet}
M.~Hu, S.~Wang, B.~Li, S.~Ning, L.~Fan, and X.~Gong, ``Penet: Towards precise and efficient image guided depth completion,'' in \emph{2021 IEEE International Conference on Robotics and Automation (ICRA)}.\hskip 1em plus 0.5em minus 0.4em\relax IEEE, 2021, pp. 13\,656--13\,662.

\bibitem{shi2023pv}
S.~Shi, L.~Jiang, J.~Deng, Z.~Wang, C.~Guo, J.~Shi, X.~Wang, and H.~Li, ``Pv-rcnn++: Point-voxel feature set abstraction with local vector representation for 3d object detection,'' \emph{International Journal of Computer Vision}, vol. 131, no.~2, pp. 531--551, 2023.

\bibitem{liu2022swin}
Z.~Liu, H.~Hu, Y.~Lin, Z.~Yao, Z.~Xie, Y.~Wei, J.~Ning, Y.~Cao, Z.~Zhang, L.~Dong \emph{et~al.}, ``Swin transformer v2: Scaling up capacity and resolution,'' in \emph{Proceedings of the IEEE/CVF conference on computer vision and pattern recognition}, 2022, pp. 12\,009--12\,019.

\bibitem{hatamizadeh2022global}
A.~Hatamizadeh, H.~Yin, J.~Kautz, and P.~Molchanov, ``Global context vision transformers,'' \emph{arXiv preprint arXiv:2206.09959}, 2022.

\bibitem{zhang2021sparse}
B.~Zhang, I.~Titov, and R.~Sennrich, ``Sparse attention with linear units,'' \emph{arXiv preprint arXiv:2104.07012}, 2021.

\bibitem{hatamizadeh2023global}
A.~Hatamizadeh, H.~Yin, G.~Heinrich, J.~Kautz, and P.~Molchanov, ``Global context vision transformers,'' 2023.

\bibitem{liu2023bevfusion}
Z.~Liu, H.~Tang, A.~Amini, X.~Yang, H.~Mao, D.~L. Rus, and S.~Han, ``Bevfusion: Multi-task multi-sensor fusion with unified bird's-eye view representation,'' in \emph{2023 IEEE International Conference on Robotics and Automation (ICRA)}.\hskip 1em plus 0.5em minus 0.4em\relax IEEE, 2023, pp. 2774--2781.

\bibitem{ye2023lidarmultinet}
D.~Ye, Z.~Zhou, W.~Chen, Y.~Xie, Y.~Wang, P.~Wang, and H.~Foroosh, ``Lidarmultinet: Towards a unified multi-task network for lidar perception,'' in \emph{Proceedings of the AAAI Conference on Artificial Intelligence}, vol.~37, no.~3, 2023, pp. 3231--3240.

\bibitem{chen2022mt}
S.~Chen, Z.~Jie, X.~Wei, and L.~Ma, ``Mt-net submission to the waymo 3d detection leaderboard,'' \emph{arXiv preprint arXiv:2207.04781}, 2022.

\bibitem{xu2022int}
J.~Xu, Z.~Miao, D.~Zhang, H.~Pan, K.~Liu, P.~Hao, J.~Zhu, Z.~Sun, H.~Li, and X.~Zhan, ``Int: Towards infinite-frames 3d detection with an efficient framework,'' in \emph{European Conference on Computer Vision}.\hskip 1em plus 0.5em minus 0.4em\relax Springer, 2022, pp. 193--209.

\bibitem{ding20201st}
Z.~Ding, Y.~Hu, R.~Ge, L.~Huang, S.~Chen, Y.~Wang, and J.~Liao, ``1st place solution for waymo open dataset challenge--3d detection and domain adaptation,'' \emph{arXiv preprint arXiv:2006.15505}, 2020.

\bibitem{zhou2022centerformer}
Z.~Zhou, X.~Zhao, Y.~Wang, P.~Wang, and H.~Foroosh, ``Centerformer: Center-based transformer for 3d object detection,'' in \emph{European Conference on Computer Vision}.\hskip 1em plus 0.5em minus 0.4em\relax Springer, 2022, pp. 496--513.

\bibitem{shi2020pv}
S.~Shi, C.~Guo, L.~Jiang, Z.~Wang, J.~Shi, X.~Wang, and H.~Li, ``Pv-rcnn: Point-voxel feature set abstraction for 3d object detection,'' in \emph{Proceedings of the IEEE/CVF Conference on Computer Vision and Pattern Recognition}, 2020, pp. 10\,529--10\,538.

\bibitem{yan2018second}
Y.~Yan, Y.~Mao, and B.~Li, ``Second: Sparsely embedded convolutional detection,'' \emph{Sensors}, vol.~18, no.~10, p. 3337, 2018.

\bibitem{li2021lidar}
Z.~Li, F.~Wang, and N.~Wang, ``Lidar r-cnn: An efficient and universal 3d object detector,'' in \emph{Proceedings of the IEEE/CVF Conference on Computer Vision and Pattern Recognition}, 2021, pp. 7546--7555.

\bibitem{mao2021pyramid}
J.~Mao, M.~Niu, H.~Bai, X.~Liang, H.~Xu, and C.~Xu, ``Pyramid r-cnn: Towards better performance and adaptability for 3d object detection,'' in \emph{Proceedings of the IEEE/CVF International Conference on Computer Vision}, 2021, pp. 2723--2732.

\bibitem{hu2022point}
J.~S. Hu, T.~Kuai, and S.~L. Waslander, ``Point density-aware voxels for lidar 3d object detection,'' in \emph{Proceedings of the IEEE/CVF Conference on Computer Vision and Pattern Recognition}, 2022, pp. 8469--8478.

\bibitem{yang2022graph}
H.~Yang, Z.~Liu, X.~Wu, W.~Wang, W.~Qian, X.~He, and D.~Cai, ``Graph r-cnn: Towards accurate 3d object detection with semantic-decorated local graph,'' in \emph{European Conference on Computer Vision}.\hskip 1em plus 0.5em minus 0.4em\relax Springer, 2022, pp. 662--679.

\bibitem{yang20213d}
Z.~Yang, Y.~Zhou, Z.~Chen, and J.~Ngiam, ``3d-man: 3d multi-frame attention network for object detection,'' in \emph{Proceedings of the IEEE/CVF Conference on Computer Vision and Pattern Recognition}, 2021, pp. 1863--1872.

\bibitem{sun2020scalability}
P.~Sun, H.~Kretzschmar, X.~Dotiwalla, A.~Chouard, V.~Patnaik, P.~Tsui, J.~Guo, Y.~Zhou, Y.~Chai, B.~Caine, V.~Vasudevan, W.~Han, J.~Ngiam, H.~Zhao, A.~Timofeev, S.~Ettinger, M.~Krivokon, A.~Gao, A.~Joshi, S.~Zhao, S.~Cheng, Y.~Zhang, J.~Shlens, Z.~Chen, and D.~Anguelov, ``Scalability in perception for autonomous driving: Waymo open dataset,'' 2020.

\bibitem{wilson2023argoverse}
B.~Wilson, W.~Qi, T.~Agarwal, J.~Lambert, J.~Singh, S.~Khandelwal, B.~Pan, R.~Kumar, A.~Hartnett, J.~K. Pontes \emph{et~al.}, ``Argoverse 2: Next generation datasets for self-driving perception and forecasting,'' \emph{arXiv preprint arXiv:2301.00493}, 2023.

\bibitem{Argoverse2}
B.~Wilson, W.~Qi, T.~Agarwal, J.~Lambert, J.~Singh, S.~Khandelwal, B.~Pan, R.~Kumar, A.~Hartnett, J.~K. Pontes, D.~Ramanan, P.~Carr, and J.~Hays, ``Argoverse 2: Next generation datasets for self-driving perception and forecasting,'' in \emph{Proceedings of the Neural Information Processing Systems Track on Datasets and Benchmarks (NeurIPS Datasets and Benchmarks 2021)}, 2021.

\bibitem{fan2022fully}
L.~Fan, F.~Wang, N.~Wang, and Z.~Zhang, ``Fully sparse 3d object detection,'' 2022.

\end{thebibliography}
\end{document}